\documentclass[10pt,twocolumn,letterpaper]{article}

\usepackage{cvpr}              
\PassOptionsToPackage{review}{cvpr}

\usepackage{multirow}
\usepackage{xcolor}
\title{ Med-2D SegNet: A Light Weight  Deep Neural Network for Medical 2D Image Segmentation}

\author{
  Lameya Sabrin \\
  \and
  Md. Sanaullah Chowdhury \\
  \and
  Salauddin Tapu \\
  \and
  Noyon Kumar Sarkar \\
  \and
  Ferdous Bin Ali \\
}

\begin{document}
\maketitle
\begin{abstract}
Accurate and efficient medical image segmentation is crucial for advancing clinical diagnostics and surgical planning, yet remains a complex challenge due to the variability in anatomical structures and the demand for low-complexity models. In this paper, we introduced Med-2D SegNet, a novel and highly efficient segmentation architecture that delivers outstanding accuracy while maintaining a minimal computational footprint. Med-2D SegNet achieves state-of-the-art performance across multiple benchmark datasets, including KVASIR-SEG, PH2, EndoVis, and GLAS, with an average Dice similarity coefficient (DSC) of 89.77\% across 20 diverse datasets. Central to its success is the compact Med Block, a specialized encoder design that incorporates dimension expansion and parameter reduction, enabling precise feature extraction while keeping model parameters to a low count of just 2.07 million. Med-2D SegNet excels in cross-dataset generalization, particularly in polyp segmentation, where it was trained on KVASIR-SEG and showed strong performance on unseen datasets, demonstrating its robustness in zero-shot learning scenarios, even though we acknowledge that further improvements are possible. With top-tier performance in both binary and multi-class segmentation, Med-2D SegNet redefines the balance between accuracy and efficiency, setting a new benchmark for medical image analysis. This work paves the way for developing accessible, high-performance diagnostic tools suitable for clinical environments and resource-constrained settings, making it a step forward in the democratization of advanced medical technology.
 
\end{abstract}    
\section{Introduction}
Segmentation is a cornerstone of biomedical image analysis, playing a vital role in identifying key regions for diagnosis, treatment, and research. In tasks like detecting cancerous cells or delineating anatomical structures, precise segmentation is indispensable for clinicians, enabling accurate and informed decision-making. However, traditional manual segmentation is both time-consuming and requires extensive domain expertise, making it impractical for widespread use. This growing challenge has fueled the demand for automated segmentation methods. 


\begin{figure}[h!] 
    \captionsetup{font=small} 
    \centering
    \includegraphics[width=\linewidth, trim=0cm 0cm 0cm 0cm, clip]{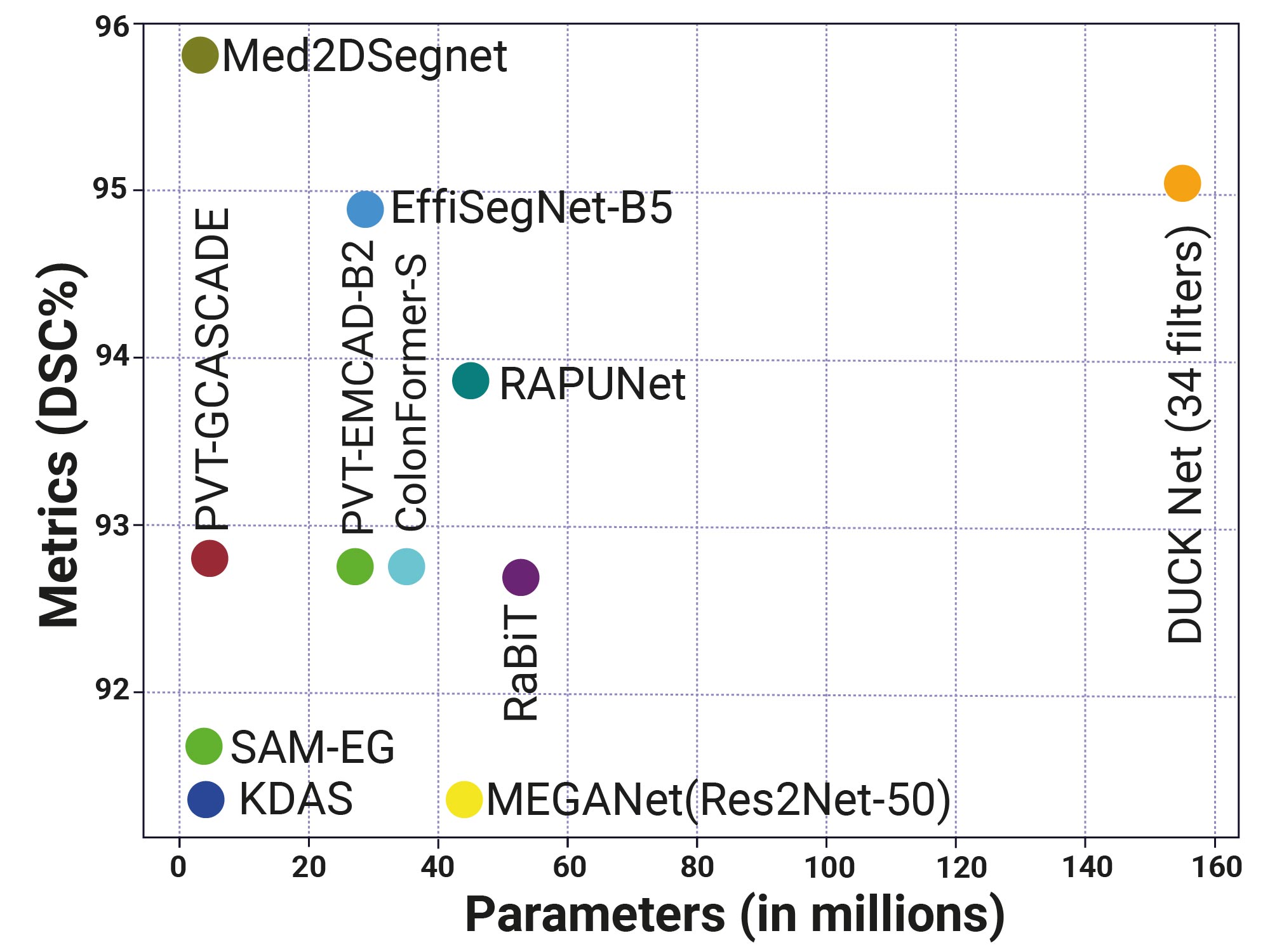}
    \caption{Comparison of model parameters vs. DSC on the KVASIR-SEG dataset. Our proposed method achieved a 95.78\% Dice score, outperforming competing models with the lowest parameter count, demonstrating an optimal balance between efficiency and segmentation accuracy.}
    \label{fig:scatter-plot}
\end{figure}

Advancements in automated segmentation have transformed tasks such as tumor detection in MRI \cite{gtifa2024multimodal} and organ segmentation in CT scans \cite{li2023eres}, demonstrating how machine learning models now tackle increasingly complex challenges with remarkable accuracy \cite{rahman2024emcad, lee2024metaformer, liu2024swin, ma2024u, rahman2024g}. Traditional CNN-based architectures like U-Net \cite{ronneberger2015u} and DeepLab \cite{chen2017deeplab} are well-regarded for their ability to extract detailed features with a reasonable computational cost, achieving widespread adoption in medical image analysis.
However, the emergence of Transformer-based models has marked a paradigm shift, particularly with Vision Transformers (ViT) \cite{thisanke2023semantic}, which excel at segmenting high-resolution images. Despite their improved accuracy, Transformers demand substantial computational resources due to their higher parameter counts, making their deployment in real-world medical imaging systems challenging. This juxtaposition between CNNs and Transformers underscores a pivotal question in segmentation research: how to leverage advanced architectures while addressing their scalability and resource-intensiveness for practical applications.
Segmentation models for multi-domain tasks often struggle with a trade-off between accuracy and computational cost. Models with large parameter counts excel in handling diverse applications but incur high resource requirements \cite{nguyen2021multi}, whereas smaller models are frequently limited to single-domain tasks \cite{galdran2022state}. To address this limitation, we proposed a lightweight segmentation model that bridges the gap between parameter efficiency and multi-domain accuracy. The proposed architecture reduces the number of parameters and weights while maintaining competitive performance across domains.
Moreover, the model's ability to adapt to varying batch sizes challenges the common notion that segmentation models are inherently batch size dependent. This adaptability not only enhances its robustness but also demonstrates its suitability for medical applications, where precision and computational efficiency are critical. By offering rapid inference without sacrificing accuracy, the proposed model balances cutting-edge performance with practical utility.

The scatter plot (Figure \ref{fig:scatter-plot}), illustrates the trade-off between model size (in millions of parameters) and segmentation accuracy (measured by the DSC \%) on the KVASIR-SEG dataset. Our model (Med-2D SegNet) achieves the highest DSC at 95.78\% with a relatively low parameter count, showcasing its efficiency and effectiveness. In contrast, other models with significantly higher parameters do not surpass this performance level, indicating that our model provides an optimal balance between accuracy and computational demands.

In this paper, we present Med-2D SegNet, a groundbreaking architecture tailored for biomedical image segmentation, which excels in capturing both intricate and high-level features. By introducing a novel design, the network reduces parameter size while significantly enhancing performance, offering a flexible configuration adaptable to varying image depths and widths. Our extensive experiments across twelve diverse domain datasets showcase its ability to achieve state-of-the-art results in most of the datasets, consistently outperforming transformer-based and CNN models, sometimes by a remarkable margin. This work lays a promising foundation for future innovations in network design, enabling more efficient modeling of long-range dependencies and advancing the field of biomedical imaging.

\section{Related Work}
Recent advancements in medical image segmentation have primarily focused on overcoming the limitations inherent to modality-specific and organ-specific models. Traditional methods such as U-Net have provided foundational architectures for segmentation tasks but face constraints like redundancy in feature integration and limited adaptability to diverse medical imaging modalities \cite{ronneberger2015u,isensee2021nnu}. Addressing these limitations, ConvNeXt \cite{liu2022convnet} introduced an inverse bottleneck block, achieving significant improvements in vision tasks by emphasizing efficient feature extraction and computational efficiency. MedNeXt \cite{roy2023mednext} adapted this concept effectively to 3D medical imaging, demonstrating superior performance in volumetric segmentation tasks by capturing rich spatial context.

Further advancements in 2D medical image segmentation have seen the development of architectures such as TransFuse \cite{zhang2021transfuse} and Swin-Unet \cite{cao2023swin}. TransFuse merges CNN and Transformer modules to leverage both local spatial details and global context, resulting in enhanced segmentation accuracy. Similarly, Swin-Unet employs a Swin Transformer backbone to efficiently capture hierarchical representations, significantly improving model generalization across various medical image segmentation tasks.

Additionally, recent architectures like M$^2$SNET \cite{zhao2023m} and M2U-Net \cite{laibacher2019m2u} propose innovative approaches addressing specific shortcomings of traditional U-shaped frameworks. M$^2$SNET introduces a subtraction unit (SU) designed to extract unique information between adjacent feature levels, effectively reducing redundancy and enhancing localization accuracy. It further integrates multi-scale processing and LossNet modules to refine segmentation quality and convergence during training. M2U-Net, on the other hand, adopts multi-level features aggregation to efficiently integrate contextual information, improving segmentation accuracy across different medical imaging modalities.
These advancements highlight ongoing efforts toward developing versatile, accurate, and computationally efficient segmentation frameworks capable of effectively handling diverse medical imaging datasets and tasks.

M2U-Net, tailored for retinal vessel segmentation, reduces the parameter count to 0.55 million while achieving state-of-the-art performance on datasets like HRF and CHASE-DB1, making it suitable for mobile applications \cite{9025339}. Overall, these models collectively represent significant advancements in the field, enhancing the accuracy and efficiency of medical image segmentation while paving the way for more adaptable and generalizable solutions in healthcare settings.

\begin{figure*}[!ht]
  \centering
  \includegraphics[width=0.9\linewidth]{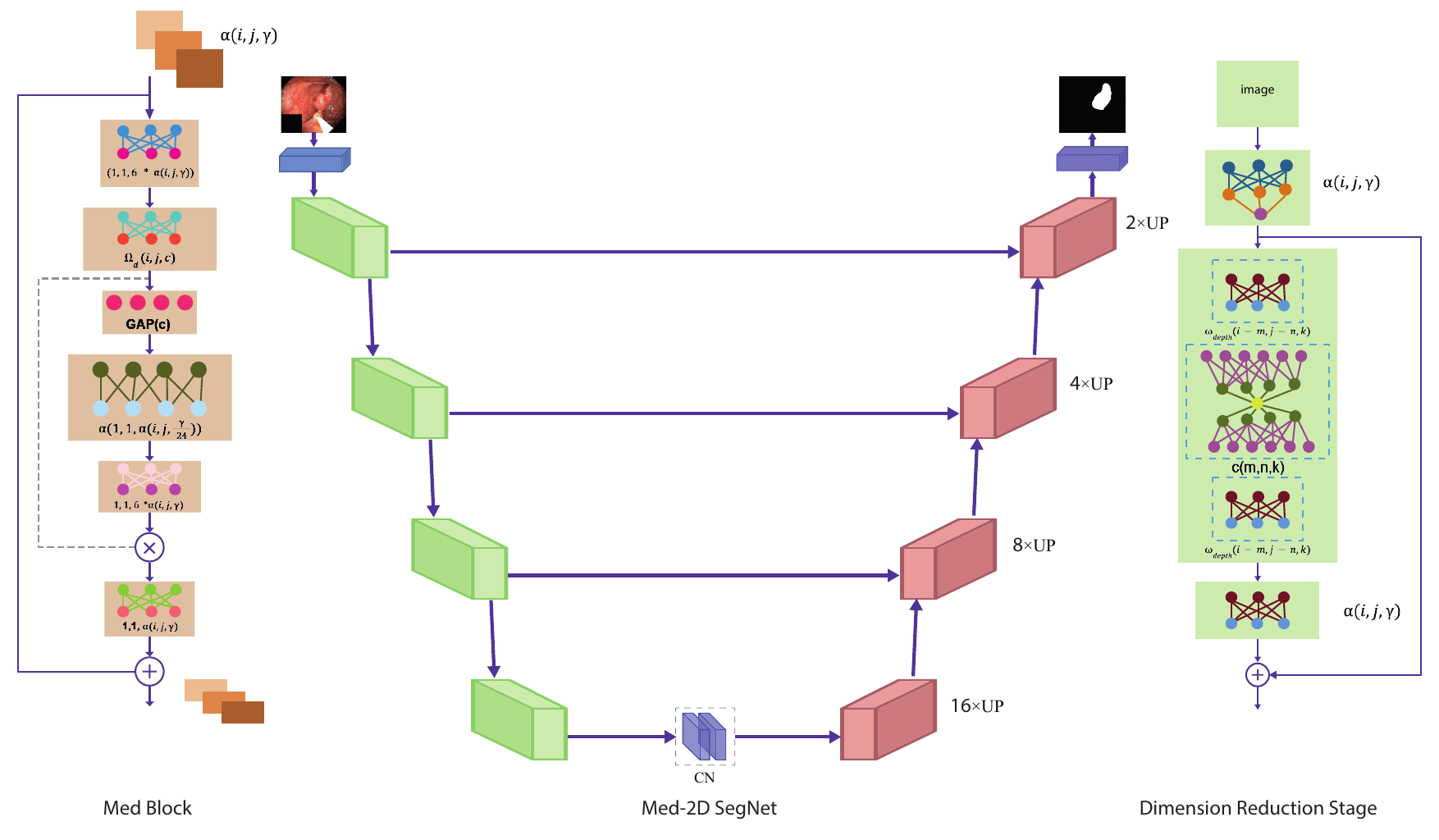}
  \caption{The middle figure illustrates the Med-2D SegNet architecture with an encoder-decoder structure. The left block represents the Med Block, detailing each layer's functionality within the main architecture. The right block illustrates the dimension reduction stages in the encoder part. In the diagram, the green blocks denote the dimensional reduction phases, where image dimensions are progressively reduced by factors of \(1/2\), \(1/4\), \(1/8\), and \(1/16\), effectively compressing the spatial information. The red blocks represent the decoder stages, which progressively reconstruct the spatial dimensions while integrating residual connections to refine the segmentation output. Together, these components form the Med-2D SegNet's encoder and decoder structure, ensuring efficient feature extraction and accurate reconstruction.}
  \label{fig:main-arc}
\end{figure*}
\section{Methodology}
The Med-2D SegNet presents a novel encoder-decoder architecture designed to efficiently recognize both simple and complex features while maintaining a compact model size with only a few million parameters. The architecture is optimized for medical image segmentation, balancing performance and computational efficiency. Figure \ref{fig:main-arc} illustrates the detailed architecture of Med-2D SegNet. In the following sections, we first introduce the building blocks of the network, followed by a comprehensive demonstration of the entire Med-2D SegNet framework.

\subsection{Med Block}
While Med-2D SegNet may visually resemble U-Net with its symmetrical encoder-decoder architecture, its underlying mechanics set it apart as a cutting-edge solution for segmentation tasks. The cornerstone of Med-2D SegNet is the Med Block, an innovative module meticulously engineered through comprehensive research and iterative optimization.  This block is designed to effectively capture intricate features across both shallow and deep layers. This architectural innovation arises from well-considered design choices, each aimed at optimizing feature extraction and enhancing overall performance. Through analysis, Med-2D SegNet showcases superior representational capacity, offering a more robust and powerful alternative to traditional segmentation models.

The Med Block enhances multi-scale feature extraction by expanding the feature map dimensionality, not merely scaling its values. Specifically, given an input feature map $\alpha \in \mathbb{R}^{H \times W \times C}$, a $1 \times 1$ convolution is applied to project it into a higher-dimensional space, resulting in an expanded feature map $\tilde{\alpha} \in \mathbb{R}^{H \times W \times 6C}$. This expansion enriches the representational capacity of the network by increasing the number of channels sixfold, thereby enabling the model to capture more diverse and complex features without altering the spatial resolution. The entire operation of the layer, combining both expansion and convolution, can be represented as:

\vspace{-.5cm}
\hspace{-.5cm}
{\[
\Omega_{ex}(i, j, \gamma) = \sum_{\alpha =1}^{C} \Psi(1 \times 1, \tilde{\alpha }(i, j, \gamma)) 
\]
\vspace{-.49cm}
\[
= \sum_{\alpha =1}^{C} \Psi(1 \times 1, 6 \cdot \alpha (i, j, \gamma)) \quad \text{for } \gamma \in \{1, 2, \ldots, 6C\}
\]
}

After expanding the feature map \( \alpha (i, j, \gamma) \) by a factor of 6, i.e., {\( \tilde{\alpha }(i, j, \gamma) = 6 \cdot \alpha (i, j, \gamma) \)}, the dimensionality explodes from \( C \) to \( 6C \), massively boosting the network’s ability to capture intricate features which basically increase the width of the layer.

.

To enhance feature extraction, we employ a depthwise convolution using a $7 \times 7$ kernel, which expands the receptive field to capture a broader range of spatial information. For a given spatial position $(i, j)$ and channel $c$, the output is expressed as:
\vspace{-.25cm}
\[
\Omega_d(i,j,c) = \sum_{p=1}^{7} \sum_{q=1}^{7} \Phi(i + p - 4, j + q - 4, c) \cdot \Psi_c(p,q)
\]

This equation describes a $7 \times 7$ convolution operation, where $p-4$ and $q-4$ center the filter over the pixel $(i,j)$, ensuring that the filter symmetrically captures neighboring pixels within the range from $(i-3,j-3)$ to $(i+3,j+3)$. The element-wise multiplication $\Phi(i + p - 4, j + q - 4, c) \cdot \Psi_c(p, q)$ allows the model to learn localized spatial features, effectively capturing patterns and spatial dependencies.

Following the Global Average Pooling (GAP) layer, which reduces the feature maps from \(H \times W \times C\) to \(1 \times 1 \times C\) by averaging:
\vspace{-.30cm}
\[
GAP(c) = \frac{1}{H \times W} \sum_{i=1}^{H} \sum_{j=1}^{W} \Omega_d(i,j,c),
\]

after reshaping the output feature map, we implement a parameter reduction layer to further streamline the model. This layer processes the input feature map \(\Phi(i,j,\alpha )\) through a \(1 \times 1\) filter \(\Psi(1,1,\alpha ,\gamma)\), generating the output \(\Omega_r(i,j,\gamma)\). The operation can be elegantly expressed as:

\vspace{-.7cm}
\[
\Omega_r(i,j,\gamma) = \sum_{\alpha =1}^{C} \Phi(i,j,\alpha ) \cdot \Psi(1,1,\alpha ,\gamma) \quad 
\]
\vspace{-.5cm}
\[
\text{for } \gamma \in \left\{1, 2, \ldots, \frac{F_1}{24}\right\}.
\]

In this equation, \(\Omega_r(i,j,\gamma)\) represents the refined output at spatial position \((i,j)\) for filter \(\gamma\), while \(\Phi(i,j,\alpha )\) denotes the input feature map for channel \(\alpha \). By strategically reducing the output channels to \(\frac{F_1}{24}\), we significantly minimize the number of parameters. This reduction enhances computational efficiency while improving the model's ability to capture intricate features, resulting in a streamlined architecture that excels in performance.

In the Med Block architecture, feature integration from the convolution and depthwise convolution layers is achieved through element-wise multiplication. Let \(\mathbf{C} \in \mathbb{R}^{H \times W \times F}\) represent the output of the convolution operation following the parameter reduction layer, computed as \(\mathbf{C}(i,j,\gamma) = \sum_{\alpha =1}^{C} \Phi(i,j,\alpha ) \cdot \Psi(1,1,\alpha ,\gamma)\), and let \(\mathbf{D} \in \mathbb{R}^{H \times W \times F}\) denote the depthwise convolution output, given by \(\mathbf{D}(i,j,\gamma) = \sum_{p=1}^{7} \sum_{q=1}^{7} \Phi(i + p - 4, j + q - 4, c) \cdot \Psi_k(p,q)\). The combined output \(\mathbf{O} \in \mathbb{R}^{H \times W \times F}\) is expressed as \(\mathbf{O}(i,j,\gamma) = \mathbf{C}(i,j,\gamma) \odot \mathbf{D}(i,j,k)\), enhancing the model's capacity to capture complex patterns.

Finally, the last convolution layer's output \(\mathbf{C}_{\text{last}}(i,j,\gamma) = \Omega(i,j,\gamma)\) is combined with the input feature map \(\mathbf{I} \in \mathbb{R}^{H \times W \times F}\) using a residual connection: \(\mathbf{O}_{\text{final}}(i,j,\gamma) = \mathbf{I}(i,j,\gamma) + \mathbf{C}_{\text{last}}(i,j,\gamma)\). This approach preserves original features and enhances gradient flow, improving overall model performance.

\subsection{Med-2D SegNet Architecture}

The Med-2D SegNet architecture is designed to optimize feature extraction through a hierarchical filter scaling mechanism, balancing computational efficiency and representational capacity. A key feature of this design is the scaling factor \( r = 1.32^2 \), which systematically adjusts the number of filters across the network. Starting with initial filter counts \( F_1 = 32 \) and \( F_2 = 24 \), subsequent filter counts are calculated iteratively as:  
\[
F_n = \text{int}(r \cdot F_{n-1}) \quad \text{for } n \in \{3, 4, \ldots, 11\}.
\]  
The scaling factor \( r \) offers key advantages by progressively reducing filter counts in deeper layers, ensuring efficient parameter distribution, and focusing computational resources on early layers for intensive feature extraction. This hierarchical approach enhances feature refinement, allowing earlier layers to capture detailed local features while deeper layers extract global patterns. Additionally, \( r \) provides a scalable, structured model design, minimizing manual tuning and achieving a balance between depth, width, and computational efficiency.

The encoder consists of three key blocks. First, the Convolution Block processes the input feature map \( X \in \mathbb{R}^{H \times W \times C} \), generating an output \( C \in \mathbb{R}^{H' \times W' \times F_1} \) via:  
\[
C(i,j,k) = \sum_{m=1}^{H} \sum_{n=1}^{W} X(m,n,k) \cdot W(i-m,j-n,k) + b(k),
\]  
where \( W \) and \( b \) represent the convolutional filters and biases, respectively. Next, the Repeated Block incorporates the Med Block, which is repeated multiple times based on the input image size to enhance depth and feature extraction capability. Following this, the Depthwise Convolution Block refines the feature maps with depthwise convolution, producing an output \( D \in \mathbb{R}^{H' \times W' \times F_1} \), defined as:  
\[
D(i,j,k) = \sum_{m=1}^{H'} \sum_{n=1}^{W'} C(m,n,k) \cdot W_{\text{depth}}(i-m,j-n,k),
\]  
where \( W_{\text{depth}} \) denotes the depthwise convolution filters. The outputs of the convolution and depthwise convolution blocks are combined via element-wise addition:  O(i,j,k) = C(i,j,k) + D(i,j,k). The decoder reconstructs high-resolution feature maps using residual blocks, which reintroduce information from earlier layers. The final output is expressed as:  
\[
O_{\text{final}}(i,j,k) = I(i,j,k) + O(i,j,k),
\]

\begin{figure*}[!ht]
  \centering
  \begin{subfigure}{0.48\linewidth}
    \centering
    \includegraphics[width=\linewidth]{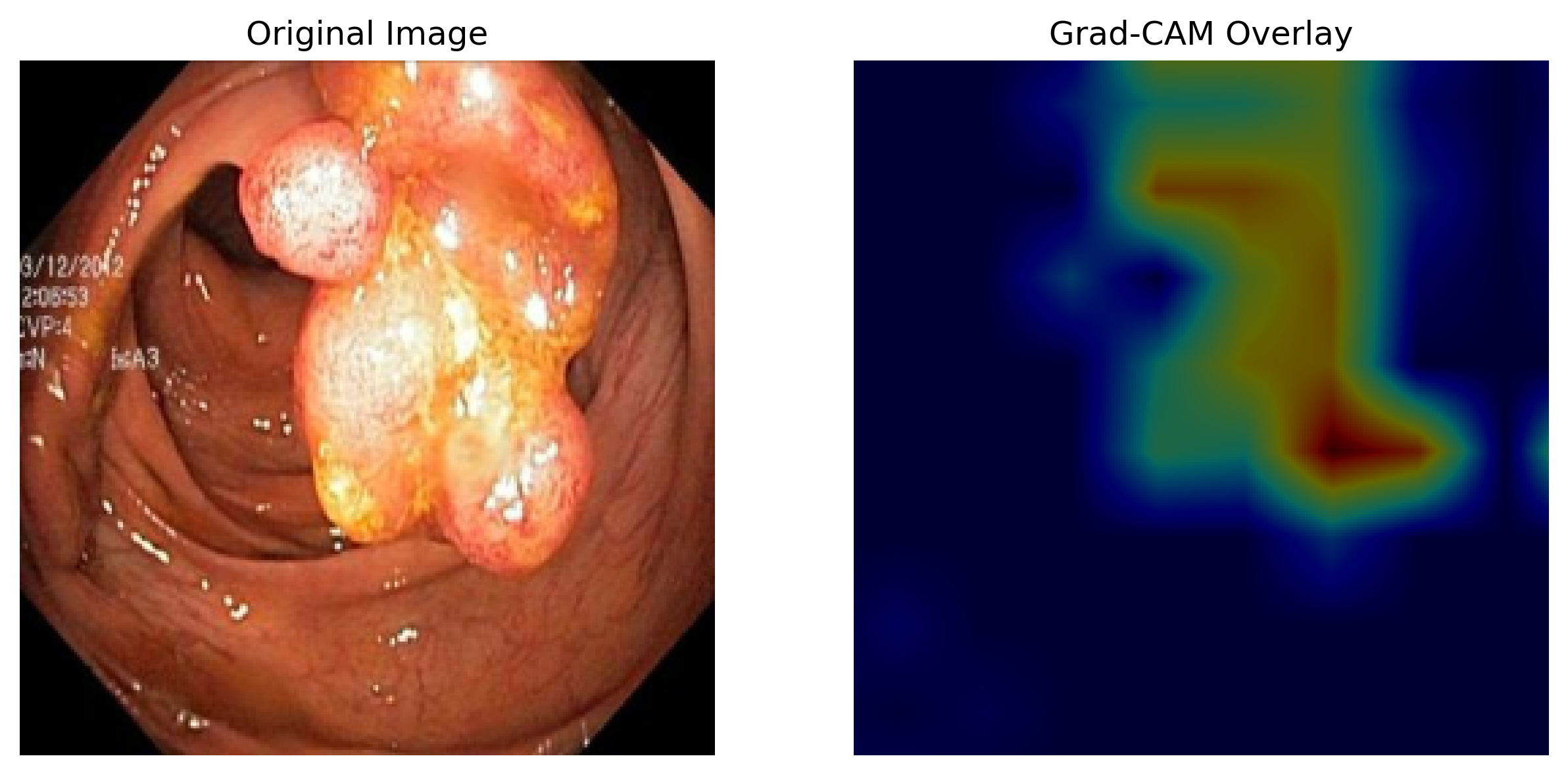}
    \caption{Grad-CAM visualizations emphasizing critical focus regions and model attention highlights.}
    \label{fig:short-c}
  \end{subfigure}
  \hfill
  \begin{subfigure}{0.48\linewidth}
    \centering
    \includegraphics[width=\linewidth]{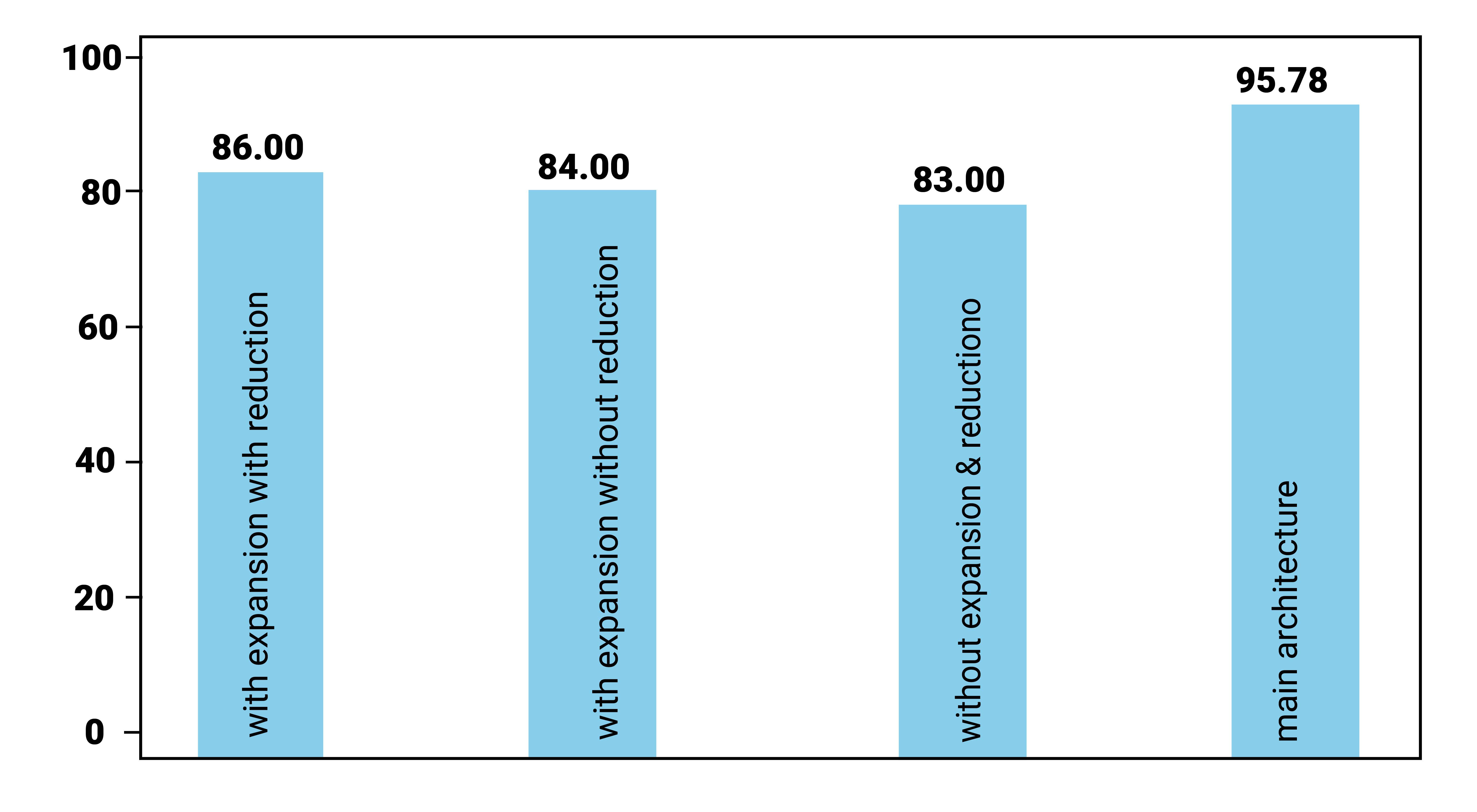}
    \caption{Comparison graph with different encoder architecture.}
    \label{fig:short-d}
  \end{subfigure}
  \caption{(a) Grad-CAM visualizations highlighting key focus areas and model attention. (b) Performance comparison across different encoder architectures, illustrating trends in model performance. Detailed analysis is provided in the Supplementary Material Parameter Complexity section.}
  \label{fig:short1}
\end{figure*}

where \( I \) is the input feature map from a previous layer. 

By leveraging the hierarchical filter scaling factor \( r \), the Med-2D SegNet achieves a compact and efficient design while maintaining the capacity to extract complex features. This makes the architecture particularly effective for image segmentation tasks, where balancing accuracy and computational cost is critical.

\section{Parameter Complexity in the Med Block Architecture}

In designing the Med Block, we aim to balance parameter complexity to enhance the model’s ability to capture detailed features while preventing overfitting and maintaining efficiency. Here, parameter complexity refers to the number of trainable parameters in the model. A high parameter count allows the model to capture intricate patterns but may lead to overfitting by learning noise instead of general features. Conversely, too few parameters may limit the model’s expressiveness, reducing its ability to capture important details needed for accurate segmentation. To strike this balance, the Med Block architecture employs two main stages: feature expansion and parameter reduction.

The Med Block begins by expanding the input feature maps to increase the model's representational power temporarily. This expansion is achieved by scaling each input feature map by a factor of 6:
\vspace{-.3cm}
\[
\tilde{\alpha }(i,j,\gamma) = 6 \cdot \alpha (i,j,\gamma)
\]

This step increases parameter complexity, allowing the model to capture detailed and nuanced patterns in the data, which is essential for high-quality segmentation.

Following feature expansion, a parameter reduction layer compresses the feature maps to a lower-dimensional representation, preventing excessive parameter growth. Mathematically, the reduction operation is expressed as:
\vspace{-.35cm}
\[
\Omega(i,j,\gamma) = \sum_{\alpha =1}^{C} \Phi(i,j,\alpha ) \cdot \Psi(1,1,\alpha ,\gamma)
\]
\[
\text{for } \gamma \in \left\{1, 2, \ldots, \frac{F_1}{24}\right\}.
\]
Here, the output channels are reduced to:
\[
F_1^{''} = \frac{F_1}{24}
\]

This reduction retains the essential features while keeping the model size manageable, avoiding overfitting, and maintaining computational efficiency. The Med Block’s combination of expansion and reduction achieves an optimal parameter complexity, balancing the expressive power of the model with the need for efficiency. The parameter count after these stages can be represented as:
Parameters after expansion: \( P_{\text{expanded}} = \sum_{j=1}^{F_1'} \text{params}(F_j) \), where \( F_1' = k \cdot F_1 \), \( k > 1 \)
Parameters after reduction: \( P_{\text{reduced}} = \sum_{j=1}^{F_1''} \text{params}(F_j) \), where \( F_1'' = \frac{F_1}{m} \), \( m > 1 \). Balancing parameter complexity in this way allows the Med Block to capture complex features necessary for accurate segmentation while keeping the model compact and computationally efficient.

By incorporating architectural variations such as expansion with and without reduction, we achieve a nuanced trade-off between model size and accuracy which is represented in Figure~\ref{fig:short-d}. These adjustments allow the model to efficiently capture relevant features while maintaining a manageable number of parameters. This balance is critical for enhancing segmentation performance, as demonstrated by the Grad-CAM visualizations Figure~\ref{fig:short-c}, which highlight the regions of interest effectively learned by the model. The design choices reflect the efficacy of our approach in leveraging parameter efficiency to achieve high performance.






\section{Experimental Setup}

\subsection{Dataset and Preprocessing}
To thoroughly assess the effectiveness and resilience of our model, we conducted an extensive series of experiments across 20 publicly available datasets which are represented in Table \ref{tab:dsc-results-on-all-data}, carefully selected for their diversity in modality, anatomical regions, and clinical targets. Notably, some datasets include 3D MRI images, which we converted into 2D slices through preprocessing steps to align with our approach. To enhance the generalization ability of the model, we applied a wide range of data augmentation techniques \cite{buslaev2020albumentations} and fine-tuned hyperparameters. Each dataset was divided into 80\% for training, 10\% for validation, and 10\% for testing. In certain cases, to ensure consistency and fairness in comparison, we adhered to the same dataset split methods as prior works; for example, DUCK-Net \cite{dumitru2023using} for polyp segmentation and U-Mamba \cite{liu2024swin} for EndoVis17. This careful design ensures that our evaluation reflects the true resilience of our model across a wide spectrum of biomedical imaging challenges. More comparisons can be found in the supplementary material Experimental Setup section.

\subsection{Implementation Details}
We resized all images to 256x256 pixels, with some datasets scaled to 512x512 as noted in the result table. Using a batch size of 128 and the Adam optimizer (learning rate: 0.0175), we trained the Med-2D SegNet with mixed activation functions—ELU and sigmoid in the encoder, ReLU in the decoder—for stronger feature extraction and decoding. Dropout regularization (rate: 0.5) was applied to prevent overfitting. Experiments with varying batch sizes (16 to 128) optimized performance, while image size variations (128 to 512) showed minimal impact. The model was trained on Kaggle’s TPU VM v3-8 with 8 accelerators and 330 GB of RAM for efficient training. More comparisons can be found in the supplementary material Experimental Setup section.

\section{Result Analysis}
In this study, we performed experiments on both binary and multiclass segmentation tasks to evaluate the stability and versatility of our model architecture. Binary segmentation aimed to isolate a specific target region, while multiclass segmentation extended this to differentiate between multiple target areas simultaneously. This approach allowed us to assess the model’s performance under varying levels of complexity and validate its capability across diverse segmentation challenges.
\subsection{Experimental Results}
Our proposed method exhibits remarkable consistency and adaptability across prominent polyp segmentation benchmarks, achieving competitive performance that underscores its robustness in diverse clinical scenarios. As displayed in Table \ref{tab:dsc-results-on-all-data}, the model attained outstanding Dice scores 0.9578 on Kvasir-SEG \cite{jha2020kvasir}, 0.9494 on Kvasir-Capsule \cite{smedsrud2021kvasir}, 0.9605 on CVC-ClinicDB \cite{bernal2015wm}, 0.9478 on CVC-ColonDB \cite{vazquez2017benchmark}, and 0.9538 on ETIS-LARIBPOLYPDB \cite{bernal2017comparative}. These consistently high scores reveal the model’s ability to handle the nuanced variations inherent to each dataset.

\begin{figure*}[!ht]
  \centering
  \begin{subfigure}{0.68\linewidth}
  \hspace{-1.2cm}
    \includegraphics[width=13cm]{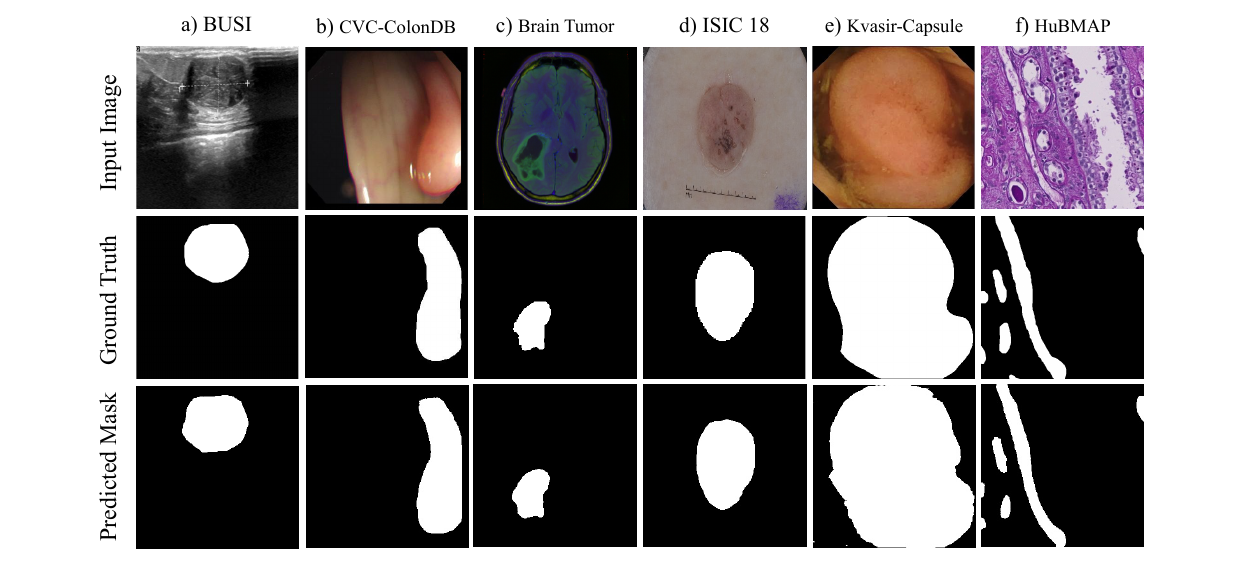}
    \caption{Predicted mask from different domain images .}
    \label{fig:short-a}
  \end{subfigure}
  \hfill
  \begin{subfigure}{0.28\linewidth}
    \hspace{-1cm}
    \includegraphics[width=7cm]{sec/Spidernks01.jpg}
    \caption{Comparison graph with other methods}
    \label{fig:short-b}
  \end{subfigure}
  \caption{ (a) Predicted results on a cross-domain dataset, demonstrating the strength and flexibility of our approach in handling diverse data. (b) Spider plot comparing the performance of our method with state-of-the-art techniques, highlighting its superior and comparable Dice score and exceptional performance across various evaluation metrics}
  \label{fig:short2}
\end{figure*}



    



\begin{table}
\centering
\caption{ Dice scores across 20 datasets spanning 12 modalities, demonstrating the model's robustness, adaptability, and outstanding performance across diverse and challenging scenarios.
}
\label{tab:dsc-results-on-all-data}
\scalebox{.7}{
\begin{tabular}{llll}
\hline
\textbf{Modality}                    & \textbf{Dataset}                                                                                       & \textbf{Image Size} & \textbf{DSC}    \\ \hline
\multirow{5}{*}{Polyp}      & Kvasir-SEG \cite{jha2020kvasir}                             & 256 x 256  & 0.9578 \\
                            & Kvasir-Capsule \cite{smedsrud2021kvasir}                    & 256 x 256  & 0.9494 \\
                            & CVC-ClinicDB \cite{bernal2015wm}                            & 256 x 256  & 0.9605 \\
                            & CVC-ColonDB \cite{vazquez2017benchmark}                     & 256 x 256  & 0.9478 \\
                            & ETIS-LARIBPOLYPDB \cite{bernal2017comparative}              & 256 x 256  & 0.9538 \\ \hline
\multirow{2}{*}{Dermoscopy} & ISIC 2018 \cite{tschandl2018ham10000}                       & 256 x 256  & 0.9128 \\
                            & PH2 \cite{6610779}                                          & 256 x 256  & 0.9641 \\ \hline
\multirow{2}{*}{Fundus}     & DRIVE \cite{asad2014ant}                                    & 512 x 512  & 0.8182 \\
                            & CHASE-DB1 \cite{fraz2012ensemble}                                   & 512 x 512  & 0.7767 \\ \hline
Retinography                & RIM One \cite{batista2020rim}                               & 128 x 128  & 0.8834 \\ \hline
Ophthalmology               & Ravir \cite{hatamizadeh2022ravir}                           & 512 x 512  & 0.7705 \\ \hline
X-Rays                      & Chest X-Ray (Lung) \cite{candemir2013lung}                  & 256 x 256  & 0.9784 \\ \hline
Ultrasound                  & Breast Ultrasound Images \cite{ALDHABYANI2020104863}        & 256 x 256  & 0.7997 \\ \hline
\multirow{4}{*}{Histology}  & CoNIC \cite{graham2024conic}                                & 256 x 256  & 0.8356 \\
                            & GLaS (Test A) \cite{sirinukunwattana2017gland}              & 256 x 256  & 0.9125 \\
                            & GLaS (Test B) \cite{sirinukunwattana2017gland}              & 256 x 256  & 0.9056 \\
                            & HuBMAP (Kidney) \cite{hubmap-hacking-the-human-vasculature} & 256 x 256  & 0.8338 \\ \hline
MRI                         & Brain Tumor \cite{Buda_2019}                               & 256 x 256  & 0.9144 \\ \hline
CT                          & LiTs \cite{bilic2023liver}                                  & 256 x 256  & 0.9202 \\ \hline
\multirow{2}{*}{Endoscopy} & Endovis 2017 (Binary) \cite{allan20192017} & 256 x 256 & 0.9307 \\
                            & Endovis 2017 (Multiclass) \cite{allan20192017}              & 256 x 256  & 0.7422 \\ \hline
Hematology                  & Cell Nuclic Segmentation (Blood) \cite{DEPTO2021101653}      & 512 x 512  & 0.9267 \\ \hline
\end{tabular}
}
\end{table}

The results presented in Table \ref{tab:dsc-results-on-all-data} demonstrate our model’s strong performance across a broad spectrum of medical imaging modalities and datasets, indicating its capability to segment diverse anatomical structures. Achieving notable Dice scores on dermoscopic (ISIC 2018: 0.9128, PH2: 0.9641), fundus (DRIVE: 0.8182), and retinography images (RIM One: 0.8834), as well as on challenging histological (GLaS Test A: 0.9125, HuBMAP: 0.8338) and MRI datasets (Brain Tumor: 0.9144), the model consistently maintains high segmentation accuracy. Furthermore, on X-ray (Chest X-Ray: 0.9784), CT scans (LiTS (binary segmentation): 0.9202), and endoscopic images (binary segmentation EndoVis 2017: 0.9307, multiclass segmentation EndoVis 2017: 0.7422), it excels in extracting detailed and clinically relevant structures. These results affirm our proposed method's generalizability and effectiveness across diverse imaging techniques, with an ability to adapt to varied tissue types and imaging resolutions—a critical feature for clinical applicability across multiple domains.

To ensure an equitable comparison, we followed the dataset splits provided by each dataset’s authors, using few-shot learning techniques to further enhance generalization. This approach reinforced the model’s capacity to maintain robust performance across datasets with distinctive characteristics. Collectively, these results highlight our model’s ability to achieve state-of-the-art segmentation accuracy across a broad range of benchmarks, emphasizing its potential as a reliable tool in real-world clinical applications.

\begin{table}[]
\caption{Results of both binary (Polyp, Skin Lesion, and Blood Vessel) and multiclass (Endovis) medical image segmentation. We reproduced the results using the publicly available implementation and data splits. Reported results of other methods are taken from their official implementation. Our method's results are shown in bold.}
\label{tab:dsc-compare}
\scalebox{.67}{
\begin{tabular}{llrr}
\hline
\textbf{Dataset} & \textbf{Method} & \multicolumn{1}{l}{\shortstack{ \textbf{Parameter} \\ \textbf{(in Millions)}}} & \multicolumn{1}{l}{\textbf{DSC}} \\ \hline 
 & PVT-EMCAD-B2 \cite{rahman2024emcad} & 26.76 & 0.9275 \\
 & DUCK-Net (34 filters) \cite{dumitru2023using} & 155.4 & 0.9502 \\
 & RAPUNet \cite{lee2024metaformer} & 43.85 & 0.9390 \\ \cline{2-4} 
\multirow{-4}{*}{Kvasir-SEG \cite{jha2020kvasir}} & \textbf{Ours} & \textbf{2.07} & \textbf{0.9578} \\ \hline
 & PVT-EMCAD-B2 \cite{rahman2024emcad} & 26.76 & 0.9521 \\
 & DUCK-Net (34 filters) \cite{dumitru2023using} & 155.4 & 0.9478 \\
 & RAPUNet \cite{lee2024metaformer} & 43.85 & 0.9610 \\ \cline{2-4} 
\multirow{-4}{*}{CVC-ClinicDB \cite{bernal2015wm}} & \textbf{Ours} & \textbf{2.07} & \textbf{0.9605} \\ \hline
 & PVT-EMCAD-B2 \cite{rahman2024emcad} & 26.76 & 0.9231 \\
 & DUCK-Net (17 filters) \cite{dumitru2023using} & 38.92 & 0.9353 \\
 & RAPUNet \cite{lee2024metaformer} & 43.85 & 0.9526 \\ \cline{2-4} 
\multirow{-4}{*}{CVC-ColonDB \cite{vazquez2017benchmark}} & \textbf{Ours} & \textbf{2.07} & \textbf{0.9478} \\ \hline
 & PVT-EMCAD-B2 \cite{rahman2024emcad} & 26.76 & 0.9229 \\
 & DUCK-Net (34 filters) \cite{dumitru2023using} & 155.4 & 0.9354 \\
 & RAPUNet \cite{lee2024metaformer} & 43.85 & 0.9572 \\ \cline{2-4} 
\multirow{-4}{*}{ETIS-LARIBPOLYPDB \cite{7840040}} & \textbf{Ours} & \textbf{2.07} & \textbf{0.9538} \\ \hline
 & PVT-GCASCADE \cite{rahman2024g} & 3.32 & 0.9151 \\
 & TransFuse \cite{zhang2021transfuse} & 26.3 & 0.9010 \\
 & UCTransNet \cite{UCTransNet} & 65.6 & 0.9050 \\ \cline{2-4} 
\multirow{-4}{*}{ISIC 2018 \cite{codella2019skin}} & \textbf{Ours} & \textbf{2.07} & \textbf{0.9128} \\ \hline
 & MERIT-CASCADE \cite{10030763} & 147.86 & 0.8221 \\
 & MERIT-GCASCADE \cite{rahman2024g} & 5.99 & 0.8290 \\
 & FR-UNet \cite{9815506} & - & 0.8316 \\ \cline{2-4} 
\multirow{-4}{*}{DRIVE \cite{asad2014ant}} & \textbf{Ours} & \textbf{2.07} & \textbf{0.8182} \\ \hline
 & nnU-Net \cite{isensee2021nnu} & 33 & 0.6264 \\
 & nnFormer \cite{zhou2023nnformer} & 60 & 0.6135 \\
 & U-Mamba\_Bot \cite{ma2024u} & 63 & 0.6540 \\
 & Swin-UMamba \cite{liu2024swin} & 28 & 0.6783 \\ \cline{2-4} 
\multirow{-5}{*}{Endovis \cite{allan20192017}} & \textbf{Ours} & \textbf{2.07} & \textbf{0.7422} \\ \hline
\end{tabular}
}
\end{table}

\subsection{Comparison with Other Methods}
Our proposed method consistently achieves high segmentation accuracy across multiple medical datasets, outperforming or matching state-of-the-art methods while maintaining a lightweight architecture. On the Kvasir-SEG dataset, we achieved a top Dice score of 0.9578, surpassing methods like DUCK-Net and RAPUNet, with far fewer parameters (2.07 million compared to DUCK-Net’s 155.4M and RAPUNet’s 43.85M). On CVC-ClinicDB, our model scored 0.9605, very close to RAPUNet's 0.961, but with a significantly smaller model size. For CVC-ColonDB and ETIS-LARIBPOLYPDB, our model achieved Dice scores of 0.9478 and 0.9538, respectively, showing similar performance to RAPUNet and DUCK-Net, but with a fraction of the parameters.

On the ISIC 2018 dataset, our model achieved a Dice score of 0.9128, closely trailing PVT-GCASCADE's 0.9151, while using a compact design, making it a strong choice for skin lesion segmentation. In blood vessel segmentation (DRIVE), we achieved a Dice score of 0.8182, and in multiclass segmentation (Endovis 2017), our model outperformed U-Mamba with a Dice score of 0.7422, compared to U-Mamba's 0.654.
With only 2.07 million parameters, our model consistently provides high Dice scores, demonstrating an optimal trade-off between accuracy and efficiency. Its compact design is especially valuable in environments with limited computational resources, offering competitive performance with minimal storage and processing overhead. This balance of accuracy and efficiency makes our model ideal for real-world medical applications. For more comparisons, see the supplementary results section.

\section{Exploring Zero-shot Learning}

Zero-shot learning (ZSL) enhances biomedical image segmentation by addressing the limitations of traditional methods that rely on large, annotated datasets, which are expensive and time-consuming to create. ZSL enables models to segment previously unseen structures or diseases, making them more effective in real-world scenarios with limited labeled data. In our study, we used a diverse training strategy using datasets such as Kvasir-Seg, DRIVE, ISIC, and PH2 to provide our model with a broad understanding of various biomedical imaging contexts. After training, we evaluated its performance on multiple test datasets, including polyp, dermatoscopic, and fundus image datasets, to assess its effectiveness.
\begin{table}[!ht]
\caption{Zero-shot learning on different benchmark datasets and comparison with other methods.}
\label{tab:zsl-compare}
 \scalebox{.77}{
\begin{tabular}{llll}
\hline
\textbf{Train Data}          & \textbf{Test Data}                 & \textbf{Method} & \textbf{DSC} \\ \hline
\multirow{10}{*}{KVASIR-SEG} & \multirow{3}{*}{CVC-ClinicDB}      & ResUNet++       & 0.6468       \\
                             &                                    & ResUNet++ + TTA & 0.6737       \\
                             &                                    & RAPUNet         & 0.8214       \\ \cline{3-4} 
                             &                                    & Ours            & 0.8319       \\ \cline{2-4} 
                             & \multirow{3}{*}{CVC-ColonDB}       & ResUNet++       & 0.5135       \\
                             &                                    & ResUNet++ + TTA & 0.5393       \\ \cline{3-4} 
                             &                                    & Ours            & 0.6877       \\ \cline{2-4} 
                             & \multirow{3}{*}{ETIS-LARIBPOLYPDB} & ResUNet++       & 0.4017       \\
                             &                                    & ResUNet++ + TTA & 0.4014       \\ \cline{3-4} 
                             &                                    & Ours            & 0.6911       \\ \hline
ISIC 2018                    & PH2                                & Ours            & 0.9397       \\ \hline
PH2                          & ISIC 2018                             & Ours            & 0.7737       \\ \hline
DRIVE                        & CHASE-DB1                          & Ours            & 0.7386       \\ \hline
\end{tabular}
}
\end{table}

The results, presented in Table \ref{tab:zsl-compare}, illustrate the performance of our zero-shot learning model compared to the established ResUNet++ framework, both with and without Test-Time Augmentation (TTA) \cite{jha2021comprehensive}, using the DSC as a key performance metric. Notably, our model achieved a DSC score of 0.8319 on the CVC-ClinicDB dataset when trained on Kvasir-Seg, significantly surpassing ResUNet++ (0.6468), ResUNet++ with TTA (0.6737) and RAPUNet (0.8214). Additionally, our model demonstrated strong zero-shot segmentation capabilities across other datasets, attaining DSC scores of 0.9397 on the PH2 dataset (trained on ISIC 2018), 0.7737 on ISIC 2018 (trained on PH2), and 0.7386 on CHASE-DB1 (trained on DRIVE).

These findings underscore the potential of zero-shot learning to revolutionize biomedical image segmentation by enabling high-quality segmentation across diverse contexts, particularly in response to emerging diseases and conditions that lack sufficient annotated data. Our results affirm that ZSL not only enhances the robustness of segmentation models but also paves the way for more efficient and scalable applications in the biomedical domain.

\section{Ablation Study}





To evaluate the impact of each model component, we conducted ablation studies by systematically disabling or replacing specific blocks. Removing the expansion block resulted in a notable drop in segmentation accuracy, emphasizing its importance in capturing complex features and improving model representation, particularly for challenging medical tasks. Omitting the parameter reduction block significantly increased the model's parameter count, raising computational demands and reducing inference efficiency, underscoring its role in maintaining compactness without sacrificing performance. Replacing our custom encoder with a standard CNN-based encoder led to a decline in overall segmentation performance, especially in multiclass tasks, highlighting the advantages of our tailored encoder design for detailed biomedical segmentation. Most of these results are visualized in Figure~\ref{fig:short-d}, with further details, including tables and additional visualizations, provided in the supplementary material’s ablation studies section.

\section{Conclusion}
In this work, we developed a CNN-based segmentation model with a custom Med Block leveraged encoder and a residual-based decoder, achieving high performance in both binary and multiclass segmentation tasks. The proposed Med Block architecture, which incorporates dimension expansion, depthwise convolution, and global average pooling, enabled a balance of computational efficiency and high performance by optimizing the parameter count without sacrificing accuracy.

In binary segmentation, the model achieved a notable 0.9307 Dice score on the EndoVis 2017 dataset, while in multiclass segmentation, it demonstrated versatility and effectiveness, handling complex 7-class tasks like segmentation of minimally invasive surgical instruments in endoscopic images and videos (Endovis 2017). To further validate adaptability, we extended our evaluation to include zero-shot performance across datasets from diverse medical fields—specifically, polyp, dermoscopic, and fundus image segmentation. This multi-domain assessment highlights our proposed method's robustness across varied imaging scenarios, underscoring its potential for broader application in medical imaging.

Future work will aim to explore additional dataset types to further stress-test our proposed method's versatility under constrained parameters. This continued refinement sets a strong foundation for our model's application in biomedical image segmentation and positions it as a promising approach for real-world, high-efficiency deployment.

{
    \small
    \bibliographystyle{ieeenat_fullname}
    \bibliography{main}
}


\end{document}